\documentclass{bmvc2k}

\usepackage{comment}
\usepackage{xcolor}

\title{Uncertainty assessment in satellite-based greenhouse gas emissions estimates using emulated atmospheric transport}
\addauthor{\hspace*{0.9cm}Jeffrey N. Clark\\[-0.2ex]\hspace*{0.9cm}{jeff.clark@bristol.ac.uk}}{}{1}
\addauthor{\hspace*{0.9cm}Elena Fillola}{}{1,2}
\addauthor{\hspace*{0.9cm}Nawid Keshtmand}{}{2}
\addauthor{\hspace*{0.9cm}Raul Santos-Rodriguez}{}{1}
\addauthor{\hspace*{0.9cm}Matthew Rigby}{}{2}

\addinstitution{
 School of Engineering Mathematics and Technology \\ University of Bristol \\
 Bristol, UK
}

\addinstitution{
 School of Chemistry \\
 University of Bristol \\
 Bristol, UK
}

\runninghead{Clark et al.}{Emulator uncertainty in emissions estimates}

\begin{document}

\maketitle

\begin{abstract}
Monitoring greenhouse gas emissions and evaluating national inventories require efficient, scalable, and reliable inference methods. Top-down approaches, combined with recent advances in satellite observations, provide new opportunities to evaluate emissions at continental and global scales. However, transport models used in these methods remain a key source of uncertainty: they are computationally expensive to run at scale, and their uncertainty is difficult to characterise. Artificial intelligence offers a dual opportunity to accelerate transport simulations and to quantify their associated uncertainty.

We present an ensemble-based pipeline for estimating atmospheric transport ``footprints'', greenhouse gas mole fraction measurements, and their uncertainties using a graph neural network emulator of a Lagrangian Particle Dispersion Model (LPDM). The approach is demonstrated with GOSAT (Greenhouse Gases Observing Satellite) observations for Brazil in 2016. The emulator achieved a $\sim$1{,}000$\times$ speed-up over the NAME LPDM, while reproducing large-scale footprint structures. Ensembles were calculated to quantify absolute and relative uncertainty, revealing spatial correlations with prediction error. The results show that ensemble spread highlights low-confidence spatial and temporal predictions for both atmospheric transport footprints and methane mole fractions.

While demonstrated here for an LPDM emulator, the approach could be applied more generally to atmospheric transport models, supporting uncertainty-aware greenhouse gas inversion systems and improving the robustness of satellite-based emissions monitoring. With further development, ensemble-based emulators could also help explore systematic LPDM errors, offering a computationally efficient pathway towards a more comprehensive uncertainty budget in greenhouse gas flux estimates.

\end{abstract}

%-------------------------------------------------------------------------
\section{Introduction}
\label{sec:intro}

Tracking of global climate commitments relies primarily on inventory based (``bottom-up'') national self-reporting of greenhouse gas (GHG) emissions. However, these approaches are increasingly complemented by top-down methods using in situ measurements and/or satellite retrievals, combined with atmospheric transport models. Atmospheric transport models are therefore an increasingly important component of national emissions evaluation systems. However, understanding of their inherent uncertainties remains a major challenge for the accurate quantification of GHG fluxes~\cite{munassar2023inverse}. Transport uncertainty arises from multiple sources, including errors in meteorological inputs, simplifications in model physics, and interpolation errors in space and time~\cite{engelen2002error}, and can affect both regional~\cite{etde_21423272, palmer2019net} and global scales~\cite{houweling2010importance, chevallier2014toward}. For example, meteorological fields have uncertainties caused by errors and gaps in observations and forecast models, and even small perturbations in wind fields can significantly alter the gas dispersion, and these errors propagate into downstream estimates of emissions~\cite{deng2017toward, liu2011co2}.

The gold standard approach to evaluating transport models has been controlled tracer-release experiments, where known emissions provide a benchmark for testing model skill~\cite{hegarty2013evaluation, simmonds2021tracers}. These physical experiments are logistically complex, spatially and temporally sparse, and expensive to perform, limiting their use for broad-scale uncertainty characterisation or to the specific locations and observed meteorological conditions, requiring computational methods to understand and quantify uncertainty. Common approaches include ensemble modelling, where perturbations in input meteorology or model parameters generate a spread of outcomes that can be used to approximate transport uncertainty \cite{delcloo2018quantification}. However, ensembles of physics simulations are particularly computationally costly for Lagrangian Particle Dispersion Models (LPDMs) run at high spatial and temporal resolutions or over extended geographical domains~\cite{lin2011studying}. These challenges are well recognised by the community, with recent policy and science roadmaps emphasising the need to better quantify transport-related uncertainty if we are to realise the full value of emerging GHG observation systems~\cite{houweling2010importance, chevallier2014toward, CEOS_CGMS_GHG_Roadmap_Issue2_2024}.

Machine learning offers attractive alternatives~\cite{eike}, and, among these, ensembles of neural networks have emerged as one of the most effective strategies for uncertainty estimation, with Bayesian Neural Networks (BNNs) \cite{blundell2015weight} and deep ensembles \cite{lakshminarayanan2017simple} often regarded as the de facto standard. Both approaches work by averaging predictions across multiple models to obtain a predictive distribution, but at the cost of substantial computational overhead. More efficient approximations of full ensembles, such as BatchEnsemble \cite{wen2020batchensemble} or Monte Carlo Dropout \cite{gal2016dropout}, have been proposed to capture similar benefits with reduced overhead, though each comes with trade-offs in accuracy or inference cost. These developments highlight the central role of ensembles in uncertainty quantification and motivate the use of machine learning emulators as a practical surrogate for computationally expensive transport models.
Recent work has demonstrated the potential of emulators to reproduce the outputs of complex transport models at significantly reduced computational cost, opening the possibility of using ensembles of such emulators as a proxy for uncertainty quantification~\cite{abdar2021review, fillola2025enabling}. If an emulator struggles to reproduce certain transport behaviours, that might provide a signal that that area or conditions lead to more inherent variability in the LPDM outputs, and therefore more uncertainty. Such approaches could provide a computationally efficient way to characterise error structure, in contrast to conventional physical ensembles.

In this paper, we take the first step towards this goal for GHG flux estimates by quantifying the emulation error of a graph neural network (GNN) LPDM emulator for GHG transport. We examine how well the ensembled emulator captures transport dynamics and use its errors as a diagnostic for uncertainty, over spatial and temporal dimensions. The pipeline and analysis that we describe here can be applied more broadly, for example to characterise the uncertainty in LPDM-only ensembles from meteorological or physical perturbations. By integrating machine learning uncertainty quantification with atmospheric transport modelling, our contribution aims to bridge the gap between computational feasibility and robust GHG emissions estimation.

%-------------------------------------------------------------------------
\begin{figure}[t]
    \centering
    \includegraphics[width=1\linewidth]{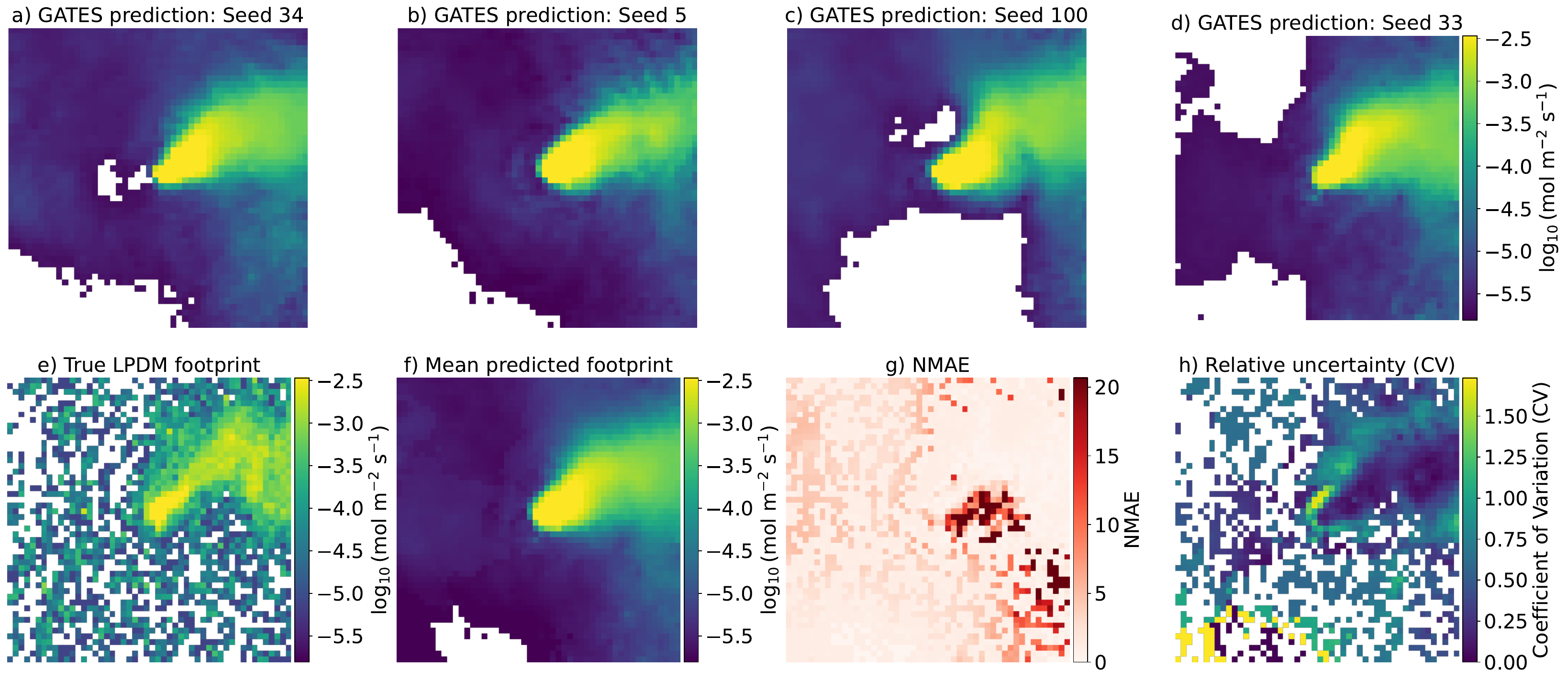}
    \caption{Top row: predicted atmospheric transport footprints generated by the four GATES (LPDM emulator) models for the same randomly selected time point from the test set. Bottom row: A comparison against the true LPDM footprint (e) with the GATES emulator ensemble mean prediction (f), normalised mean absolute error (NMAE) between the two (g), and coefficient of variation (CV) of predictions (h).}
    \label{fig:single_fps}
\end{figure}

\section{Methods}
\label{sec:methods}

\subsection{Dataset}
We utilise a GNN model to emulate LPDM-derived transport ``footprints'' (Figure~\ref{fig:single_fps}). Each footprint represents the sensitivity of a satellite measurement to surface emissions, computed by releasing thousands of hypothetical air parcels backwards in time for 30 days from the satellite sounding location (measurement point) and altitude using atmospheric state estimates. These parcels record their surface contact, yielding a two-dimensional sensitivity field. In our experiments, this occurs on a regular latitude--longitude grid ($\sim$33 $\times$ 25 km resolution) spanning 60.98$^\circ$ S to 22.32$^\circ$ N and 91.33$^\circ$ W to 24.8$^\circ$ W over South America, generated with the UK Met Office’s NAME LPDM. Footprint values are log-transformed.

The GNN model utilises 160 input features per grid cell, following recently established methods~\cite{fillola2025enabling}. The model utilises a range of time-varying meteorological features derived from the Met Office's Unified
Model (UM) global analysis fields, and static features to provide location-specific context. The meteorological features are extracted at seven vertical levels (100 m to 18 km) and at three time steps relative to the observation (0, --6, and --12 hours).  

The dataset is split by observation period as follows, training set: 2014--2015 ($\sim$11,165 footprints); validation set: Jan--Mar 2016 ($\sim$4,314 footprints); and a test set: Apr--Dec 2016 ($\sim$16,945 footprints). This separation ensures that validation and testing contain unseen meteorological conditions, preventing temporal leakage from the training set.

\subsection{Model architecture and training}

We emulate LPDM-generated footprints using the GATES framework~\cite{fillola2025enabling}, a GNN designed to replicate atmospheric transport footprints at a fraction of the computational cost. In the training phase, the model operates on a $50 \times 50$ grid centred on each observation; for the purposes of this uncertainty study inference is performed over the same $50 \times 50$ grid, with out-of-domain areas filled with zeros.  

The GATES model follows an encoder--processor--decoder structure:

\begin{itemize}
    \item \textbf{Encoder:} Maps grid inputs into an abstract triangular mesh. Local features are aggregated using multi-layer perceptrons (MLPs).
    \item \textbf{Processor:} Performs multiple rounds of message passing \cite{battaglia_relational_2018} across mesh nodes, each connected to six neighbours nodes.
    \item \textbf{Decoder:} Maps mesh features back to the original grid, predicting footprint values per cell.
\end{itemize}

Four GATES models were independently trained with different random seeds with which to initialise model weights. An ensemble of four models was chosen as the minimum viable quantity to demonstrate the technique. Training of each model takes $\sim$10 hours on a 32~GB NVIDIA V100 GPU. Post-processing steps include thresholding near-zero values to reduce noise and applying bias correction via quantile mapping using the validation set. Predictive performance was calculated as the error by subtracting the LPDM footprint values from the mean predicted values.

\subsection{Mole fraction calculations}

Each footprint (GATES-emulated or LPDM-generated) is convolved with a flux field to obtain above-baseline column methane mole fractions. Results are presented using two different methane flux fields, both re-gridded to the footprint resolution: a bottom-up map, and a uniform map. The bottom-up flux field, the same used in \cite{tunnicliffe2020quantifying} and \cite{fillola2025enabling}, aggregates anthropogenic emissions (EDGAR v4.3.2 database \cite{janssens-maenhout_edgar_2019}), biomass burning (GFED v4.1 \cite{van_der_werf_global_2017}) and wetlands (SWAMPS \cite{schroeder_development_2015}) We use the map for June 2016 throughout, to remove seasonal differences in the comparison. The uniform flux emissions field is scaled to the median magnitude of the bottom-up emissions flux field to result in interpretable results. The two flux fields therefore serve complementary roles: the bottom-up field reflects realistic spatial emissions variability, while the uniform field enables clearer attribution of uncertainty to transport processes alone.

\subsection{Uncertainty analysis}
Uncertainty was quantified for the ensemble of four GATES models two ways: firstly absolute uncertainty was considered by calculating the standard deviation of predictions, secondly relative uncertainty was measured by calculating the coefficient of variation ($\mathrm{CV}_{\text{pred}}$) as presented in Equation~\eqref{eq:cv} -- an established metric for atmospheric transport uncertainty~\cite{miller2015biases}.

\begin{equation}
    \mathrm{CV}_{\text{pred}} = \frac{\sigma_{\text{pred}}}{\mu_{\text{pred}} + \epsilon}
    \label{eq:cv}
\end{equation}
where $\sigma_{\text{pred}}$ is the standard deviation across model predictions, $\mu_{\text{pred}}$ is the predicted mean, and $\epsilon$ is a small constant for numerical stability.

These metrics are applied to both footprints and methane mole fractions, and across spatial and temporal dimensions.

%-------------------------------------------------------------------------
\section{Results}
\label{sec:results}
\subsection{Model training}

\begin{figure}
    \centering
    \includegraphics[width=1\linewidth]{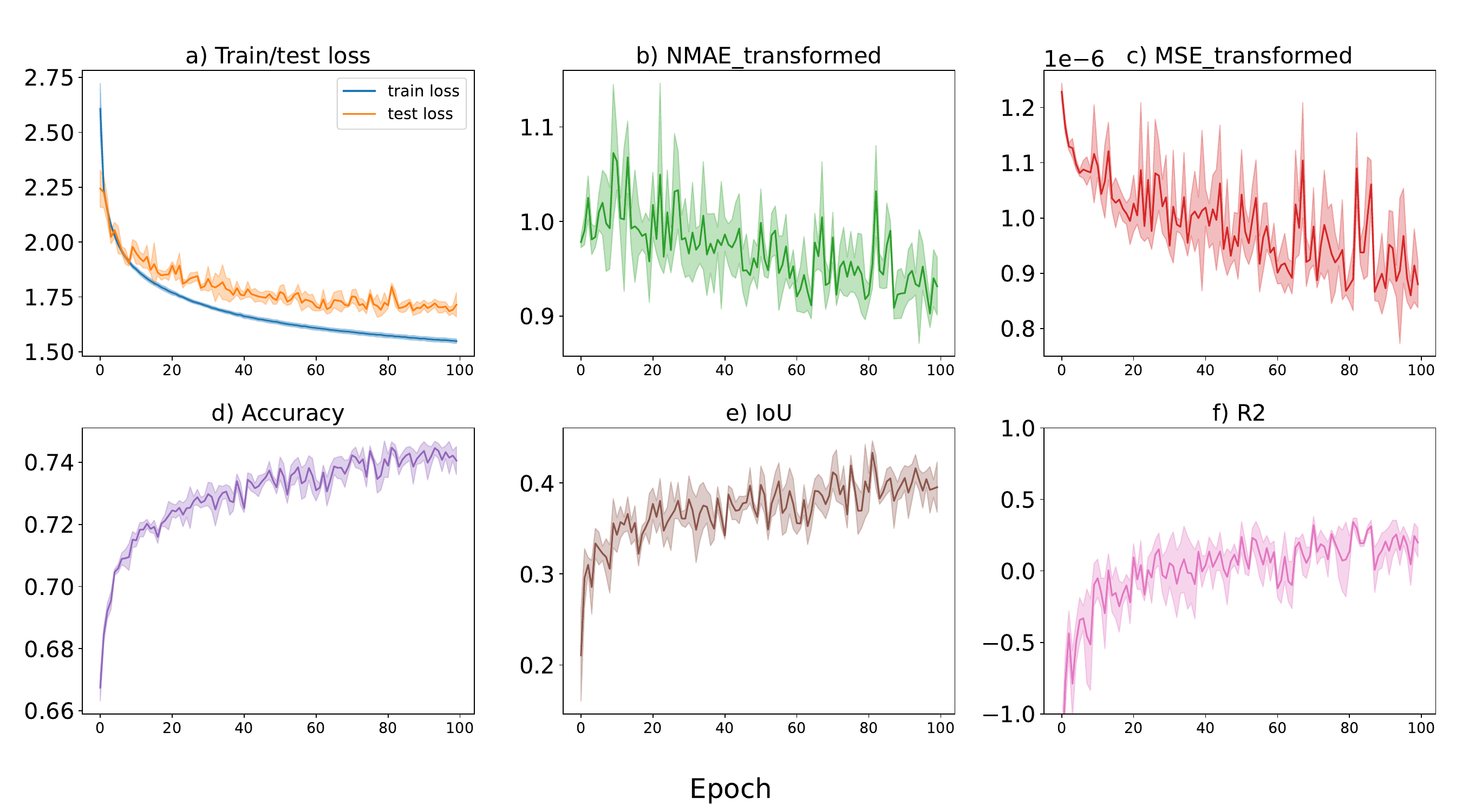}
    \caption{Mean performance during GATES LPDM emulator model development across standard machine learning metrics. Panels b-f present performance on the test set. NMAE = Normalised mean absolute error. MSE = Mean squared error. IoU = Intersection over union. R$^2$ = coefficient of determination. Error bars represent the standard deviation across the four trained models.}
    \label{fig:training_metrics}
\end{figure}

Uncertainty during model development is demonstrated across 100 training epochs for the four trained models (Figure \ref{fig:training_metrics}). Train and test loss (a) decreased steadily with limited separation between models, while both NMAE (b) and MSE (c) exhibited wider error margins, suggesting greater sensitivity to random model initialisation. Accuracy (d) improved gradually with low uncertainty bands, whereas IoU (e) and R² (f) converged with uncertainty bands approximately comparable to the variability per epoch within runs. Inference time was $\sim$0.75~s per footprint, yielding a $\sim$1{,}000$\times$ speed-up compared with a single LPDM simulation ($\sim$20~min), enabling large-scale ensemble application.

\subsection{Footprints}

Predicted footprints reproduced the large-scale structure of LPDM sensitivities(Figure~\ref{fig:single_fps}a-d, f),  capturing the plume extent and orientation of the LPDM footprint (Figure~\ref{fig:single_fps}e).

Predicted footprints from the four models broadly matched the true footprint's overall structure and each produced qualitatively similar footprints (Figure \ref{fig:single_fps}a-d), indicating a broadly consistent prediction in shape and direction of spatial sensitivity. Relative errors and inter-model uncertainty were most apparent at footprint edges and in regions of low sensitivity to surface fluxes (low footprint values) outside of the main footprint, highlighting unstable predictions (Figure~\ref{fig:single_fps}g,h).

\begin{figure}[t]
    \centering
    \includegraphics[width=0.8\linewidth]{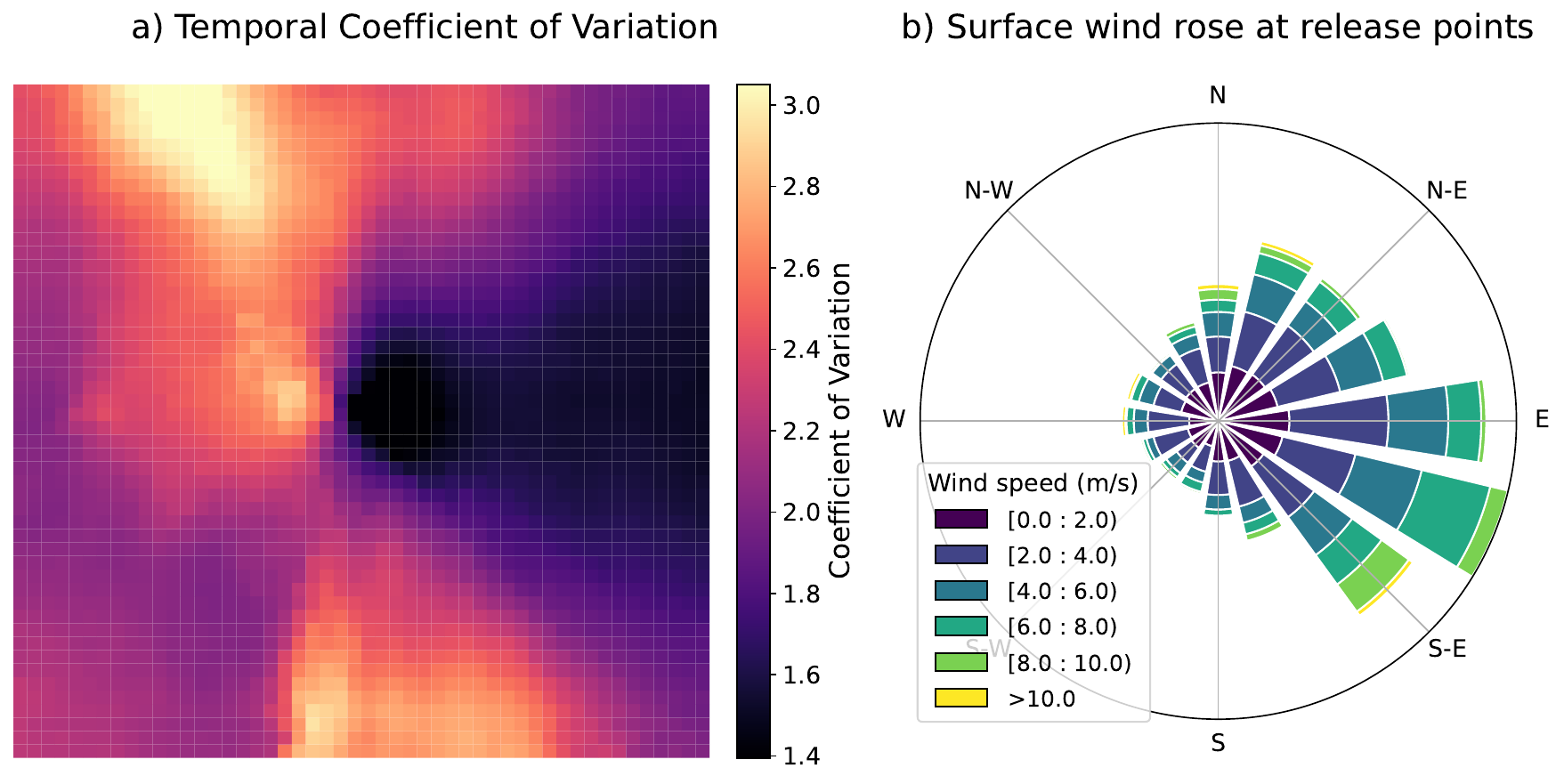}
    \caption{Left: temporal coefficient of variation of the mean prediction over the entire test set. Right: wind rose for surface-level winds, centred around the release point per footprint of the test set.}
    \label{fig:cv_windrose}
\end{figure}

\begin{figure}[t]
    \centering
    \includegraphics[width=0.98\linewidth]{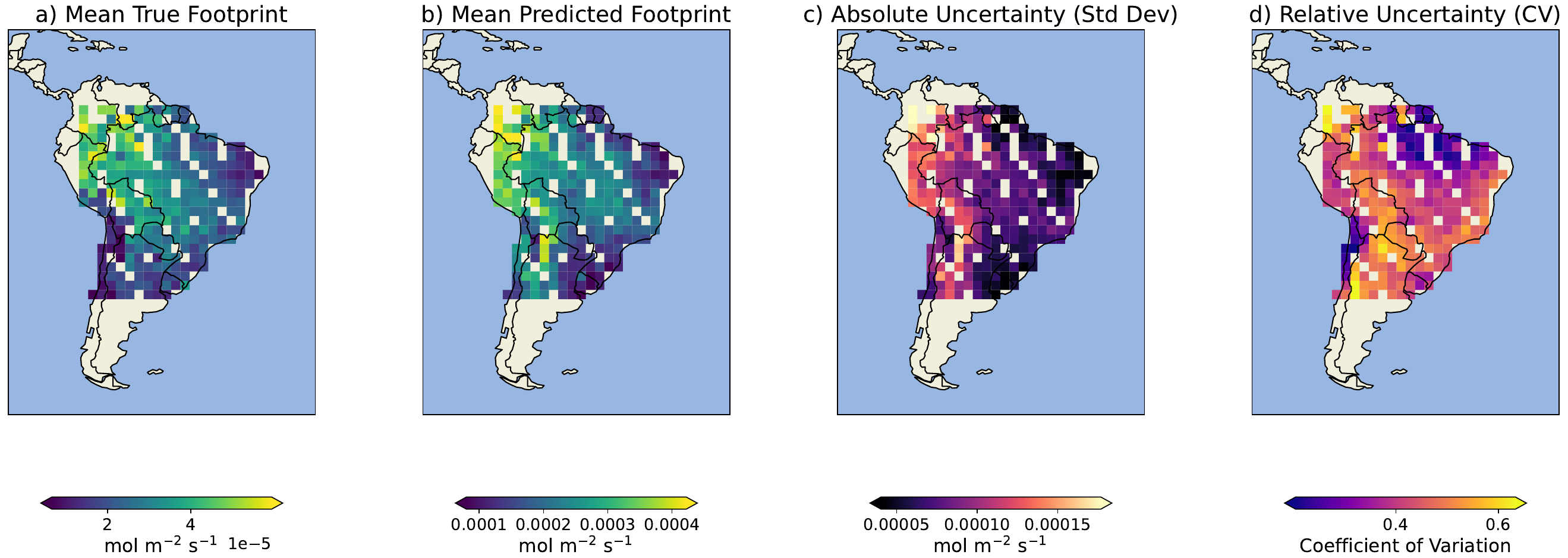}
    \caption{Spatial maps of mean footprint sensitivities across South America. a) LPDM-generated, b) GATES-predicted, c) standard deviation of predictions, and d) coefficient of variation across the four models.}
    \label{fig:footprint_uncertainty_map}
\end{figure}

Aggregated temporal and spatial uncertainties revealed structured patterns (Figures~\ref{fig:cv_windrose}, \ref{fig:footprint_uncertainty_map}). On average, per footprint, relative uncertainties were lowest in the east (Figure~\ref{fig:cv_windrose}a), aligning with persistent easterly winds (Figure~\ref{fig:cv_windrose}b). At the continent scale, lowest average uncertainties were found in north eastern South America (Figure~\ref{fig:footprint_uncertainty_map}c,d). Higher uncertainties occurred in the western regions, such as in the Andes which have more heterogeneous topography, suggesting that dynamically complex meteorological regimes reduce emulator robustness. The coefficient of variation (Figure~\ref{fig:footprint_uncertainty_map}d) further highlighted regions of low sensitivity (Figure~\ref{fig:footprint_uncertainty_map}a,b) but disproportionately high uncertainty, such as the eastern coast of the continent.

\subsection{Methane mole fraction predictions}
\begin{figure}
    \centering
    \includegraphics[width=0.9\linewidth]{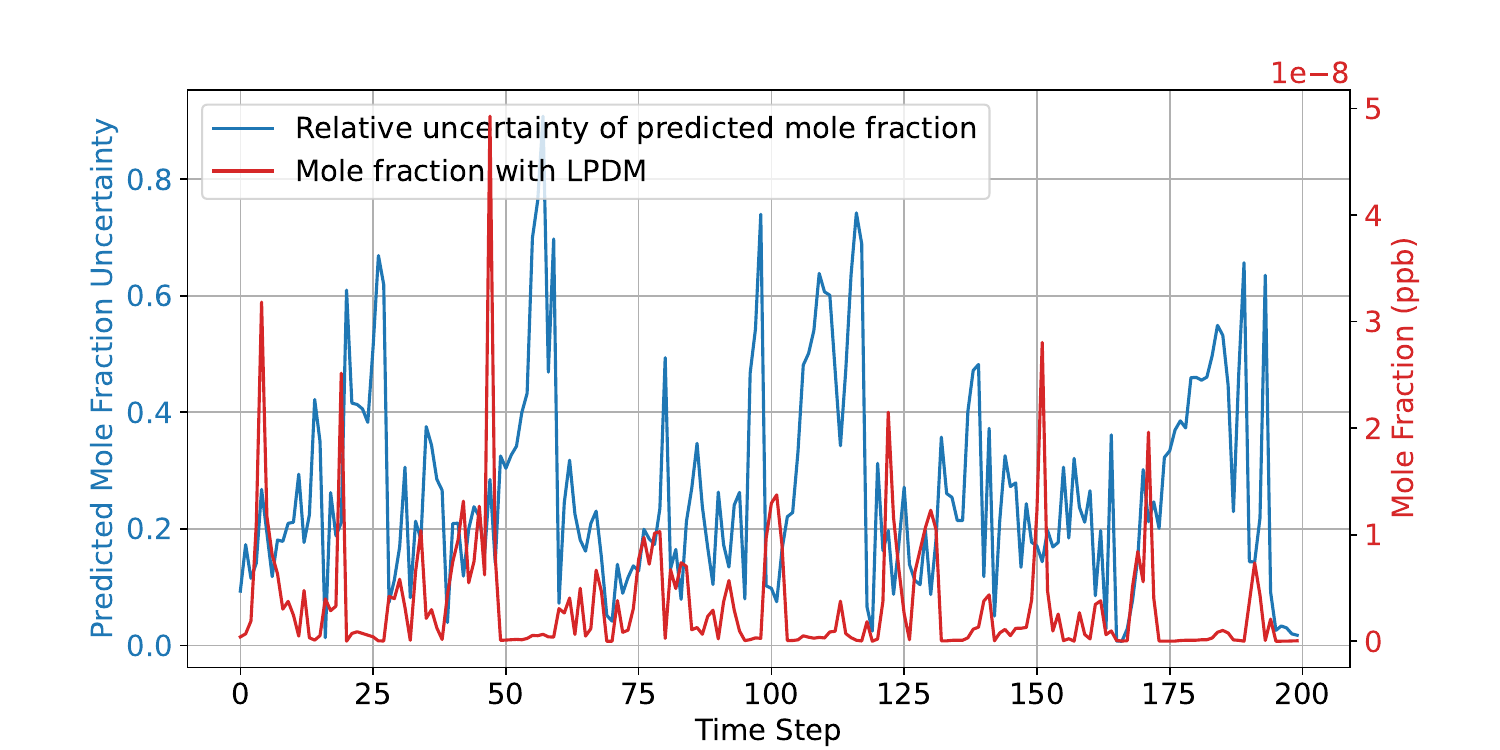}
    \caption{Relative uncertainty for predicted mole fractions across the first 200 time points of the test set, compared against mole fractions using LPDM footprints. Both are calculated using the bottom-up flux field.}
    \label{fig:uniform_mole_timeseries}
\end{figure}

The presented pipeline enables quantification of the mole fraction uncertainty resulting from the footprint emulation error. Timeseries analysis (Figure~\ref{fig:uniform_mole_timeseries}) showed temporal periods of heightened uncertainty between GNN models (blue line). Similarly large temporal fluctuations in mole fraction derived using LPDM footprints occur (red line). 

\begin{figure}[t]
\centering
% tighten table spacing so images align cleanly
\setlength{\tabcolsep}{0pt}
\renewcommand{\arraystretch}{1}

\begin{tabular}{@{}c@{}}
  \bmvaHangBox{\includegraphics[width=.95\linewidth]{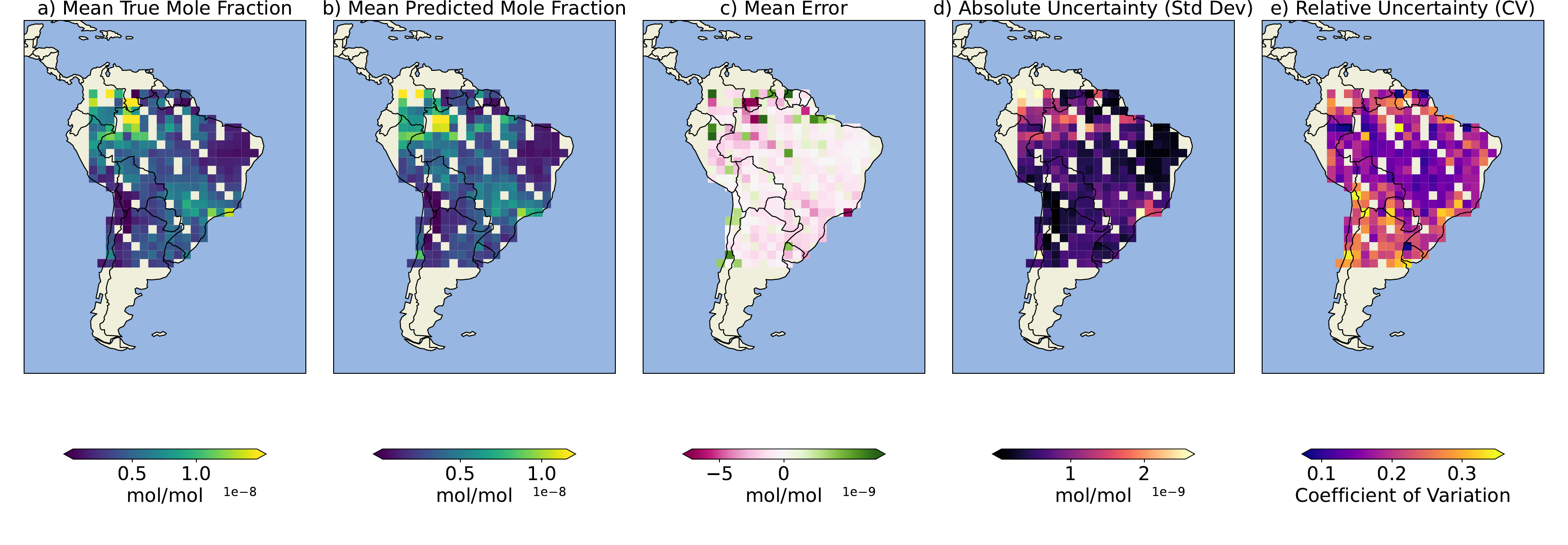}}\\[-0.25em]
  (1) With bottom-up emissions flux\\[0.75em]
  \bmvaHangBox{\includegraphics[width=.95\linewidth]{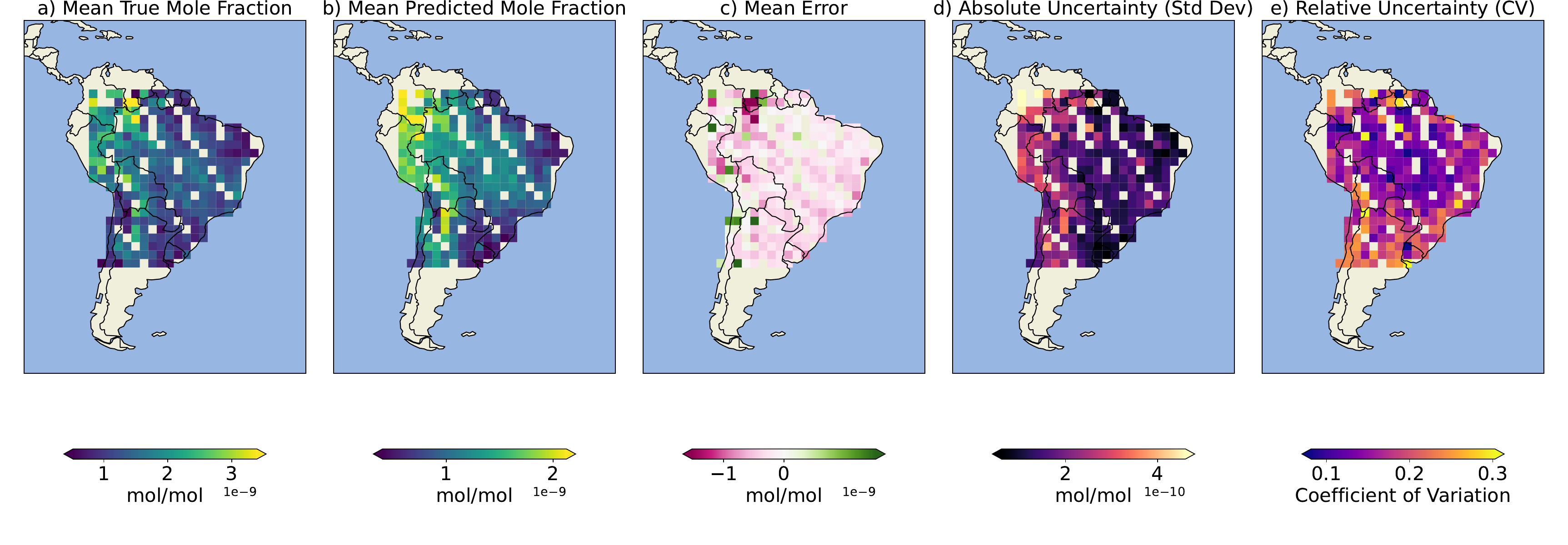}}\\[-0.25em]
  (2) With uniform flux
\end{tabular}

\caption{Spatial map of mole fractions and uncertainties with bottom-up emissions flux (1) and uniform flux (2). For each: a) true mole fractions using LPDM-based footprints, b) predicted mole fractions using GATES, c) mean error between the two, d) absolute uncertainty (standard deviation), e) relative uncertainty (coefficient of variation).}
\label{fig:mole_fraction_map_combined}
\end{figure}

While the raw time series is dense, spatial maps provided clearer insight (Figure~\ref{fig:mole_fraction_map_combined}). Panel 1 of (Figure~\ref{fig:mole_fraction_map_combined}) demonstrates that mole fraction uncertainties derived with bottom-up emissions flux broadly mirrored the footprint uncertainty structure (Figure~\ref{fig:footprint_uncertainty_map}), with higher spread in the north-west. Absolute uncertainty was largest where transport sensitivities and mole fractions were highest (Figure~\ref{fig:mole_fraction_map_combined} upper panel, plots a,b,d), while the coefficient of variation revealed relative instability in regions of low baseline sensitivity (panel e). Importantly, the spatial pattern of absolute ensemble uncertainty (panel d) qualitatively matches the mean error (panel c), indicating that uncertainty estimates are informative of prediction reliability: regions with higher spread generally coincide with larger deviations from the LPDM-derived mole fractions.

The uniform flux case (lower panel of Figure~\ref{fig:mole_fraction_map_combined}) provides a baseline in which spatial variability arises solely from transport rather than emission heterogeneity. Crucially, ensemble spread continues to align with mean error, confirming that the emulator’s uncertainty estimates capture transport-driven variability rather than artefacts of the flux field.

%-------------------------------------------------------------------------
\section{Discussion}
\label{sec:discussion}

This study demonstrates that ensembles of graph neural network transport emulators can provide both fast and informative estimates of uncertainty in atmospheric footprints and derived mole fractions. The correspondence between ensemble spread and prediction error suggests that model disagreement can serve as a practical indicator of low-confidence predictions. This finding is consistent with other deep learning domains where ensemble spread aligns with true error patterns~\cite{abdar2021review}.

Two main insights emerge. First, uncertainty is structured rather than random: it is lowest in regions of persistent easterly flow and highest in complex regions such as the Andes mountains and southern latitudes, following established findings~\cite{miller2015biases}. Second, relative uncertainty (coefficient of variance, CV) complements absolute metrics by identifying unstable predictions in low-sensitivity regions that may otherwise be overlooked.

From an applications perspective, these results have direct relevance for top-down emissions estimation. Current inversion frameworks often assume transport errors are uniform or uncorrelated~\cite{tunnicliffe2020quantifying}, whereas our approach provides data-driven identification of more uncertain emulated footprints. Incorporating these uncertainty layers could improve inversion robustness and guide prioritisation of satellite retrievals.

Several limitations remain. The ensemble size was small ($n=4$) and whilst this proved sufficient to demonstrate that ensemble spread captures regions of low predictive confidence, scaling to larger ensembles (e.g.\ 10--20 members) would provide more stable estimates. Exploring trade-offs between ensemble size, computational cost, and predictive gains remains an important direction for future work. Our demonstration focused on GOSAT methane retrievals over Brazil in 2016. Although this choice provided a well-constrained test case, further validation is required across different regions, time periods, and gas species. Inversion systems typically combine atmospheric transport, prior fluxes, and satellite retrievals within a Bayesian framework to produce optimised surface fluxes. Our current analysis isolates uncertainty associated with atmospheric transport, as represented by NAME and the GNN emulators. Other possible sources of error within the wider inversion system were not considered. Propagating and combining the multiple sources of uncertainty within a unified inversion framework is a remaining opportunity. Exploration of additional techniques for uncertainty quantification including Bayesian Neural Networks~\cite{Goan2020} or calibration~\cite{telmo} are warranted. Future work should also explore correlations between ensemble spread and input feature sparsity, systematic model errors, and inversion experiments to quantify the influence of these factors on derived fluxes.

%-------------------------------------------------------------------------
\section{Conclusions}
\label{sec:conclusions}

We introduce an ensemble-based pipeline for quantifying uncertainty in graph neural network emulators of atmospheric transport footprints. Applied to GOSAT methane retrievals over Brazil, the method revealed structured spatial and temporal uncertainty patterns, with ensemble spread reliably flagging regions and times of low predictive confidence. The approach offers actionable uncertainty estimates using a computationally efficient method. Ensemble spread correlates with emulation error, allowing selective down-weighting of uncertain predictions, using an emulator offering a $\sim$1{,}000$\times$ speed-up compared with NAME LPDM, enabling fast and scalable ensemble-scale analyses. Although demonstrated for an LPDM emulator, the framework could be applied to other atmospheric transport models. By bridging computational feasibility with robust uncertainty quantification, this work supports more reliable satellite-based greenhouse gas monitoring, directly informing climate policy and inventory verification.

%------------------------------------------------------------------------
\section*{Acknowledgements}
We thank the Met Office, for permitting the usage of the NAME model to generate the footprints, and the Unified Model to
extract the meteorology. We thank Rob Parker and the University of Leicester team for providing GOSAT satellite XCH4 retrievals.
The development and training of models, and all the analysis shown here, were carried out using the computational facilities of
the Advanced Computing Research Centre, University of Bristol.
This work was supported by UK Research and Innovation grant NE/Z504294/1 (JNC, NK, MR), a Google PhD Fellowship, 2021 (EF), and Turing AI Fellowship grant EP/V024817/1 (RSR).

\bibliography{egbib}

\begin{thebibliography}{28}
\providecommand{\natexlab}[1]{#1}
\providecommand{\url}[1]{\texttt{#1}}
\expandafter\ifx\csname urlstyle\endcsname\relax
  \providecommand{\doi}[1]{doi: #1}\else
  \providecommand{\doi}{doi: \begingroup \urlstyle{rm}\Url}\fi

\bibitem[Abdar et~al.(2021)Abdar, Pourpanah, Hussain, Rezazadegan, Liu, Ghavamzadeh, Fieguth, Cao, Khosravi, Acharya, et~al.]{abdar2021review}
Moloud Abdar, Farhad Pourpanah, Sadiq Hussain, Dana Rezazadegan, Li~Liu, Mohammad Ghavamzadeh, Paul Fieguth, Xiaochun Cao, Abbas Khosravi, U~Rajendra Acharya, et~al.
\newblock A review of uncertainty quantification in deep learning: Techniques, applications and challenges.
\newblock \emph{Information fusion}, 76:\penalty0 243--297, 2021.

\bibitem[Battaglia et~al.(2018)Battaglia, Hamrick, Bapst, Sanchez-Gonzalez, Zambaldi, Malinowski, Tacchetti, Raposo, Santoro, Faulkner, Gülçehre, Song, Ballard, Gilmer, Dahl, Vaswani, Allen, Nash, Langston, Dyer, Heess, Wierstra, Kohli, Botvinick, Vinyals, Li, and Pascanu]{battaglia_relational_2018}
Peter~W. Battaglia, Jessica~B. Hamrick, Victor Bapst, Alvaro Sanchez-Gonzalez, Vinícius~Flores Zambaldi, Mateusz Malinowski, Andrea Tacchetti, David Raposo, Adam Santoro, Ryan Faulkner, Çaglar Gülçehre, H.~Francis Song, Andrew~J. Ballard, Justin Gilmer, George~E. Dahl, Ashish Vaswani, Kelsey~R. Allen, Charles Nash, Victoria Langston, Chris Dyer, Nicolas Heess, Daan Wierstra, Pushmeet Kohli, Matthew~M. Botvinick, Oriol Vinyals, Yujia Li, and Razvan Pascanu.
\newblock Relational inductive biases, deep learning, and graph networks.
\newblock \emph{CoRR}, abs/1806.01261, 2018.
\newblock \doi{arXiv:1806.01261}.
\newblock URL \url{http://arxiv.org/abs/1806.01261}.
\newblock arXiv: 1806.01261.

\bibitem[Bergamaschi et~al.(2011)Bergamaschi, Corazza, Segers, Vermeulen, Manning, Athanassiadou, Thompson, Pison, Bousquet, and Karstens]{etde_21423272}
P~Bergamaschi, M~Corazza, A~Segers, A~Vermeulen, A~Manning, M~Athanassiadou, R~Thompson, I~Pison, P~Bousquet, and U~Karstens.
\newblock Top-down estimates of european ch4 and n2o emissions based on 5 different inverse models, April 2011.
\newblock Netherlands.

\bibitem[Blundell et~al.(2015)Blundell, Cornebise, Kavukcuoglu, and Wierstra]{blundell2015weight}
Charles Blundell, Julien Cornebise, Koray Kavukcuoglu, and Daan Wierstra.
\newblock Weight uncertainty in neural network.
\newblock In \emph{International Conference on Machine Learning}, pages 1613--1622. PMLR, 2015.

\bibitem[Chevallier et~al.(2014)Chevallier, Palmer, Feng, Boesch, O'Dell, and Bousquet]{chevallier2014toward}
Fr{\'e}d{\'e}ric Chevallier, Paul~I Palmer, Liang Feng, Hartmut Boesch, Christopher~W O'Dell, and Philippe Bousquet.
\newblock Toward robust and consistent regional co2 flux estimates from in situ and spaceborne measurements of atmospheric co2.
\newblock \emph{Geophysical Research Letters}, 41\penalty0 (3):\penalty0 1065--1070, 2014.

\bibitem[{Committee on Earth Observation Satellites (CEOS) and Coordination Group for Meteorological Satellites (CGMS)}(2024)]{CEOS_CGMS_GHG_Roadmap_Issue2_2024}
{Committee on Earth Observation Satellites (CEOS) and Coordination Group for Meteorological Satellites (CGMS)}.
\newblock {CEOS-CGMS Roadmap for a coordinated implementation of carbox dioxide and methane monitoring from space, Issue 2, v1.0}.
\newblock Technical Report Issue 2, v1.0, CEOS-CGMS Working Group on Climate, October 2024.
\newblock URL \url{https://ceos.org/document_management/Publications/Publications-and-Key-Documents/Atmosphere/CEOS_CGMS_GHG_Roadmap_Issue_2_V1.0_FINAL.pdf}.
\newblock Endorsed by CEOS Plenary-38 in 2024.

\bibitem[Delcloo and De~Meutter(2018)]{delcloo2018quantification}
Andy Delcloo and Pieter De~Meutter.
\newblock Quantification of uncertainty in lagrangian dispersion modelling, using ecmwf’s new era5 ensemble.
\newblock In \emph{International Technical Meeting on Air Pollution Modelling and its Application}, pages 343--346. Springer, 2018.

\bibitem[Deng et~al.(2017)Deng, Lauvaux, Davis, Gaudet, Miles, Richardson, Wu, Sarmiento, Hardesty, Bonin, et~al.]{deng2017toward}
Aijun Deng, Thomas Lauvaux, Kenneth~J Davis, Brian~J Gaudet, Natasha Miles, Scott~J Richardson, Kai Wu, Daniel~P Sarmiento, R~Michael Hardesty, Timothy~A Bonin, et~al.
\newblock Toward reduced transport errors in a high resolution urban co2 inversion system.
\newblock \emph{Elem Sci Anth}, 5:\penalty0 20, 2017.

\bibitem[Engelen et~al.(2002)Engelen, Denning, and Gurney]{engelen2002error}
Richard~J Engelen, A~Scott Denning, and Kevin~R Gurney.
\newblock On error estimation in atmospheric co2 inversions.
\newblock \emph{Journal of Geophysical Research: Atmospheres}, 107\penalty0 (D22):\penalty0 ACL--10, 2002.

\bibitem[Fillola et~al.(2025)Fillola, Santos-Rodriguez, Tunnicliffe, Clark, Keshtmand, Ganesan, and Rigby]{fillola2025enabling}
Elena Fillola, Raul Santos-Rodriguez, Rachel Tunnicliffe, Jeffrey Clark, Nawid Keshtmand, Anita Ganesan, and Matthew Rigby.
\newblock Enabling fast greenhouse gas emissions inference from satellites with gates: a graph-neural-network atmospheric transport emulation system.
\newblock Under review with EGUsphere, 2025.

\bibitem[Gal and Ghahramani(2016)]{gal2016dropout}
Yarin Gal and Zoubin Ghahramani.
\newblock Dropout as a bayesian approximation: Representing model uncertainty in deep learning.
\newblock In \emph{international conference on machine learning}, pages 1050--1059. PMLR, 2016.

\bibitem[Goan and Fookes(2020)]{Goan2020}
Ethan Goan and Clinton Fookes.
\newblock \emph{Bayesian Neural Networks: An Introduction and Survey}, pages 45--87.
\newblock Springer International Publishing, Cham, 2020.

\bibitem[Hegarty et~al.(2013)Hegarty, Draxler, Stein, Brioude, Mountain, Eluszkiewicz, Nehrkorn, Ngan, and Andrews]{hegarty2013evaluation}
Jennifer Hegarty, Roland~R Draxler, Ariel~F Stein, Jerome Brioude, Marikate Mountain, Janusz Eluszkiewicz, Thomas Nehrkorn, Fong Ngan, and Arlyn Andrews.
\newblock Evaluation of lagrangian particle dispersion models with measurements from controlled tracer releases.
\newblock \emph{Journal of Applied Meteorology and Climatology}, 52\penalty0 (12):\penalty0 2623--2637, 2013.

\bibitem[Houweling et~al.(2010)Houweling, Aben, Breon, Chevallier, Deutscher, Engelen, Gerbig, Griffith, Hungershoefer, Macatangay, et~al.]{houweling2010importance}
S~Houweling, I~Aben, F-M Breon, F~Chevallier, Nicholas Deutscher, R~Engelen, C~Gerbig, David Griffith, K~Hungershoefer, Ronald Macatangay, et~al.
\newblock The importance of transport model uncertainties for the estimation of co 2 sources and sinks using satellite measurements.
\newblock \emph{Atmospheric chemistry and physics}, 10\penalty0 (20):\penalty0 9981--9992, 2010.

\bibitem[{Huellermeier, Eyke and Waegeman, Willem}({2021})]{eike}
{Huellermeier, Eyke and Waegeman, Willem}.
\newblock {Aleatoric and epistemic uncertainty in machine learning : an introduction to concepts and methods}.
\newblock \emph{{MACHINE LEARNING}}, {110}\penalty0 ({3}):\penalty0 {457--506}, {2021}.
\newblock ISSN {0885-6125}.
\newblock URL \url{{http://doi.org/10.1007/s10994-021-05946-3}}.

\bibitem[Janssens-Maenhout et~al.(2019)Janssens-Maenhout, Crippa, Guizzardi, Muntean, Schaaf, Dentener, Bergamaschi, Pagliari, Olivier, Peters, van Aardenne, Monni, Doering, Petrescu, Solazzo, and Oreggioni]{janssens-maenhout_edgar_2019}
G.~Janssens-Maenhout, M.~Crippa, D.~Guizzardi, M.~Muntean, E.~Schaaf, F.~Dentener, P.~Bergamaschi, V.~Pagliari, J.~G.~J. Olivier, J.~A. H.~W. Peters, J.~A. van Aardenne, S.~Monni, U.~Doering, A.~M.~R. Petrescu, E.~Solazzo, and G.~D. Oreggioni.
\newblock {EDGAR} v4.3.2 {Global} {Atlas} of the three major greenhouse gas emissions for the period 1970–2012.
\newblock \emph{Earth System Science Data}, 11\penalty0 (3):\penalty0 959--1002, 2019.
\newblock \doi{10.5194/essd-11-959-2019}.
\newblock URL \url{https://essd.copernicus.org/articles/11/959/2019/}.

\bibitem[Lakshminarayanan et~al.(2017)Lakshminarayanan, Pritzel, and Blundell]{lakshminarayanan2017simple}
Balaji Lakshminarayanan, Alexander Pritzel, and Charles Blundell.
\newblock Simple and scalable predictive uncertainty estimation using deep ensembles.
\newblock \emph{Advances in neural information processing systems}, 30, 2017.

\bibitem[Lin et~al.(2011)Lin, Brunner, and Gerbig]{lin2011studying}
John~C Lin, Dominik Brunner, and Christoph Gerbig.
\newblock Studying atmospheric transport through lagrangian models.
\newblock \emph{Eos, Transactions American Geophysical Union}, 92\penalty0 (21):\penalty0 177--178, 2011.

\bibitem[Liu et~al.(2011)Liu, Fung, Kalnay, and Kang]{liu2011co2}
Junjie Liu, Inez Fung, Eugenia Kalnay, and Ji-Sun Kang.
\newblock Co2 transport uncertainties from the uncertainties in meteorological fields.
\newblock \emph{Geophysical Research Letters}, 38\penalty0 (12), 2011.

\bibitem[Miller et~al.(2015)Miller, Hayek, Andrews, Fung, and Liu]{miller2015biases}
SM~Miller, MN~Hayek, AE~Andrews, I~Fung, and J~Liu.
\newblock Biases in atmospheric co 2 estimates from correlated meteorology modeling errors.
\newblock \emph{Atmospheric Chemistry and Physics}, 15\penalty0 (5):\penalty0 2903--2914, 2015.

\bibitem[Munassar et~al.(2023)Munassar, Monteil, Scholze, Karstens, R{\"o}denbeck, Koch, Totsche, and Gerbig]{munassar2023inverse}
Saqr Munassar, Guillaume Monteil, Marko Scholze, Ute Karstens, Christian R{\"o}denbeck, Frank-Thomas Koch, Kai~U Totsche, and Christoph Gerbig.
\newblock Why do inverse models disagree? a case study with two european co 2 inversions.
\newblock \emph{Atmospheric Chemistry and Physics}, 23\penalty0 (4):\penalty0 2813--2828, 2023.

\bibitem[Palmer et~al.(2019)Palmer, Feng, Baker, Chevallier, B{\"o}sch, and Somkuti]{palmer2019net}
Paul~I Palmer, Liang Feng, David Baker, Fr{\'e}d{\'e}ric Chevallier, Hartmut B{\"o}sch, and Peter Somkuti.
\newblock Net carbon emissions from african biosphere dominate pan-tropical atmospheric co2 signal.
\newblock \emph{Nature communications}, 10\penalty0 (1):\penalty0 3344, 2019.

\bibitem[Schroeder et~al.(2015)Schroeder, McDonald, Chapman, Jensen, Podest, Tessler, Bohn, and Zimmermann]{schroeder_development_2015}
Ronny Schroeder, Kyle~C. McDonald, Bruce~D. Chapman, Katherine Jensen, Erika Podest, Zachary~D. Tessler, Theodore~J. Bohn, and Reiner Zimmermann.
\newblock Development and {Evaluation} of a {Multi}-{Year} {Fractional} {Surface} {Water} {Data} {Set} {Derived} from {Active}/{Passive} {Microwave} {Remote} {Sensing} {Data}.
\newblock \emph{Remote Sensing}, 7\penalty0 (12):\penalty0 16688--16732, 2015.
\newblock ISSN 2072-4292.
\newblock \doi{10.3390/rs71215843}.
\newblock URL \url{https://www.mdpi.com/2072-4292/7/12/15843}.

\bibitem[Silva~Filho et~al.(2023)Silva~Filho, Song, Perello-Nieto, Santos-Rodriguez, Kull, and Flach]{telmo}
Telmo Silva~Filho, Hao Song, Miquel Perello-Nieto, Raul Santos-Rodriguez, Meelis Kull, and Peter Flach.
\newblock Classifier calibration: a survey on how to assess and improve predicted class probabilities.
\newblock \emph{Machine Learning}, 112\penalty0 (9):\penalty0 3211--3260, 2023.

\bibitem[Simmonds et~al.(2021)Simmonds, Palmer, Rigby, McCulloch, O'Doherty, and Manning]{simmonds2021tracers}
PG~Simmonds, PI~Palmer, M~Rigby, A~McCulloch, S~O'Doherty, and AJ~Manning.
\newblock Tracers for evaluating computational models of atmospheric transport and oxidation at regional to global scales.
\newblock \emph{Atmospheric environment}, 246:\penalty0 118074, 2021.

\bibitem[Tunnicliffe et~al.(2020)Tunnicliffe, Ganesan, Parker, Boesch, Gedney, Poulter, Zhang, Lavri{\v{c}}, Walter, Rigby, et~al.]{tunnicliffe2020quantifying}
Rachel~L Tunnicliffe, Anita~L Ganesan, Robert~J Parker, Hartmut Boesch, Nicola Gedney, Benjamin Poulter, Zhen Zhang, Jo{\v{s}}t~V Lavri{\v{c}}, David Walter, Matthew Rigby, et~al.
\newblock Quantifying sources of brazil's ch 4 emissions between 2010 and 2018 from satellite data.
\newblock \emph{Atmospheric Chemistry and Physics}, 20\penalty0 (21):\penalty0 13041--13067, 2020.

\bibitem[van~der Werf et~al.(2017)van~der Werf, Randerson, Giglio, van Leeuwen, Chen, Rogers, Mu, van Marle, Morton, Collatz, Yokelson, and Kasibhatla]{van_der_werf_global_2017}
G.~R. van~der Werf, J.~T. Randerson, L.~Giglio, T.~T. van Leeuwen, Y.~Chen, B.~M. Rogers, M.~Mu, M.~J.~E. van Marle, D.~C. Morton, G.~J. Collatz, R.~J. Yokelson, and P.~S. Kasibhatla.
\newblock Global fire emissions estimates during 1997–2016.
\newblock \emph{Earth System Science Data}, 9\penalty0 (2):\penalty0 697--720, 2017.
\newblock \doi{10.5194/essd-9-697-2017}.
\newblock URL \url{https://essd.copernicus.org/articles/9/697/2017/}.

\bibitem[Wen et~al.(2020)Wen, Tran, and Ba]{wen2020batchensemble}
Yeming Wen, Dustin Tran, and Jimmy Ba.
\newblock Batchensemble: an alternative approach to efficient ensemble and lifelong learning.
\newblock \emph{arXiv preprint arXiv:2002.06715}, 2020.

\end{thebibliography}
\end{document}